\documentclass[letterpaper,10 pt,journal]{IEEEtran}
\usepackage{amsmath,amsfonts}
\usepackage{algorithmic}
\usepackage{algorithm}
\usepackage{array}
\usepackage[caption=false,font=normalsize,labelfont=sf,textfont=sf]{subfig}
\usepackage{textcomp}
\usepackage{stfloats}
\usepackage{url}
\usepackage{verbatim}
\usepackage{graphicx}
\usepackage{cite}
\usepackage{multirow} 
\usepackage{booktabs}
\title{\LARGE \bf
Breaking of brightness consistency in optical flow with a lightweight CNN network
}

\author{Yicheng Lin$^1$, Shuo Wang$^1$, Yunlong Jiang and Bin Han, \textit{Member, IEEE}% <-this % stops a space
\thanks{$^1$These authors contributed equally to this work.}% <-this % stops a space
\thanks{This work was supported in part by the National Natural Science Foundation of China (52375015) and in part by the Natural Science Foundation of Hubei Province of China (2022CFB239). (Corresponding author: Bin Han)}% <-this % stops a space
\thanks{Y. Lin, S. Wang, Y. Jiang and B. Han are with the State Key Laboratory of Intelligent Manufacturing Equipment and Technology, School of Mechanical Science and Engineering, Huazhong University of Science and Technology, Wuhan 430074, China (e-mail: {\{\tt\small yichenglin,shuowang99,jiangyunlong binhan \}@hust.edu.cn}).}%
}

\begin{document}
\thispagestyle{empty}
\pagestyle{empty}
\maketitle

\begin{abstract}
% 稀疏光流法是计算机视觉基础任务之一，被广泛应用于视觉里程计，运动检测中。然而稀疏光流算法假设环境的亮度是恒定的，这限制了其在高动态场景（HDR）下的应用。本文采用数据驱动的方法学习图像中具有光照不变性的特征图。在特征图不变的假设下，实现不依赖于亮度恒定假设的光流法。具体来说，一个轻量级网络被用于提取图像中的光照不变的卷积特征以及角。通过特征图构建特征金字塔并对所提取的角进行光流跟踪。所提出的网络仅采用4次卷积操作，在板载CPU上帧率可达190fps。端到端的无监督方法被用于训练网络。由于浅层网络难以获得可靠的角，因此一个额外的深层网络被用于辅助the shallow one training. 为了验证所提出的方法，在动态光照环境下进行了特征跟踪实验，并替换了VINS-Mono中的光流法得到更准确的视觉惯性系统。The code are publicly available at https://github.com/linyicheng1/LET-NET.
The sparse optical flow method is a fundamental task in computer vision. However, its reliance on the assumption of constant environmental brightness constrains its applicability in high dynamic range (HDR) scenes. In this study, we propose a novel approach aimed at transcending image color information by learning a feature map that is robust to illumination changes. This feature map is subsequently structured into a feature pyramid and integrated into sparse Lucas-Kanade (LK) optical flow. By adopting this hybrid optical flow method, we circumvent the limitation imposed by the brightness constant assumption. Specifically, we utilize a lightweight network to extract both the feature map and keypoints from the image. Given the challenge of obtaining reliable keypoints for the shallow network, we employ an additional deep network to support the training process. Both networks are trained using unsupervised methods. The proposed lightweight network achieves a remarkable speed of 190fps on the onboard CPU. To validate our approach, we conduct comparisons of repeatability and matching performance with conventional optical flow methods under dynamic illumination conditions. Furthermore, we demonstrate the effectiveness of our method by integrating it into VINS-Mono, resulting in a significantly reduced translation error of 93\% on a public HDR dataset. The code implementation is publicly available at https://github.com/linyicheng1/LET-NET.
\end{abstract}


\section{Introduction}

% 光流法是值连续图像之间的像素移动距离估计的算法,包括稀疏光流法和稠密光流法
% 稀疏光流法只估计关键点在图像之间的像素移动距离,而稠密光流法则估计所有位置对应的像素移动距离
% 光流估计是计算机视觉的经典问题，它被广泛应用在目标跟踪，运动检测，视觉里程计等领域。
% 例如,VINS-MOno和OV2SLAM采用稀疏光流法作为视觉SLAM中关键的模块之一
% 光流法在动态光照下的性能决定了这些算法对光照的鲁棒性.
\IEEEPARstart{O}{ptical} flow refers to the pixels motion between consecutive frames in an image sequence and optical flow method is an algorithm for estimating pixel motion between consecutive images, which includes sparse optical flow method and dense optical flow method. Sparse optical flow method only estimates the pixel motion distance of keypoints, while dense optical flow method estimates in all positions. Optical flow estimation is a classic problem in computer vision. For instance, VINS-Mono\cite{vins-mono} and OV$^2$SLAM \cite{ov2} utilize sparse optical flow method as a key module in visual Simultaneous Localization and Mapping (vSLAM). The performance of optical flow method in HDR scenes determines their robustness to illumination changes. Therefore, the study of illumination-robust optical flow methods is crucial.\par 

% 传统的稀疏光流方法发展过程中，普遍融入了环境光照恒定的假设，这限制了其在高动态范围（HDR）场景中的表现。尽管后续[8,37]等人提出梯度一致性和其他高阶一致性约束，如the constancy of the Hessian and the constancy of the Laplacian，能够提升光照鲁棒性。却只是缓解了光照敏感问题。另一方面，基于学习的稠密光流方法,FlowNet2.0,虽然可以克服这一问题，但其效率较低，通常需要GPU支持，不太适用于CPU实时应用。综上所述，目前尚无已知在CPU上能够实现实时运行且不依赖静态光照假设的稀疏光流方法。这一现状导致当前最先进的视觉SLAM算法仍然倾向于选择传统的稀疏光流，而在HDR场景中却难以维持其稳健性。
The traditional sparse optical flow methods \cite{lk, plk} have historically assumed static environmental illumination, which limits their effectiveness in HDR scenarios. Although subsequent research \cite{papenberg2006highly} have introduced gradient consistency and other higher-order constraints like the constancy of the Hessian and Laplacian to improve illumination robustness, these approaches only partially address this issue. Learning-based dense optical flow methods, such as FlowNet \cite{flownet}, offer a solution to this problem. However, they tend to be less efficient and often require GPU support. Thus, there is no sparse optical flow method capable of real-time execution on CPUs without relying on prior assumptions. Consequently, even the most advanced vSLAM \cite{ov2} still struggle to maintain robustness in HDR scenes. \par 

% 我们采集了不同方向灯光下的图片，展示出所提出方法对光照的鲁棒性。其中，forward optical flow 指在第一张图片中提取角点并在第二张图片中跟踪。而backward optical flow 则是交换两张图片的处理。
\begin{figure}[!t]
        \centering
        \includegraphics[width=0.48\textwidth]{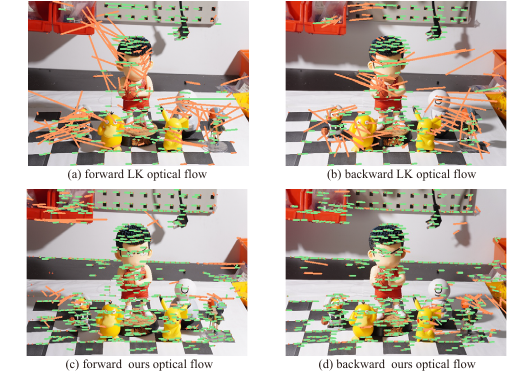}
        \caption{\textbf{Examples of dynamic lighting scene images.} We collected images under different directions of light to demonstrate the robustness of the proposed method to illumination. Among them, forward optical flow refers to extracting keypoints in the first image and tracking them in the second image. The backward optical flow is the opposite.} 
        \label{fig_i1} 
        \end{figure} 

% 在本工作中，通过结合深度学习方法对光照的鲁棒性和传统方法的高效性，混合光流的概念被首次提出。
% 一个轻量的卷积网络被用于同时提取卷积特征F和角点得分图S。卷积特征是超越RGB原本图像的高级特征，具有一定的光照不变性与视角不变性。具有强不变性的图像点位被提取作为角点，并认为这些点位能够在光流中更有效的跟踪。在两帧图像中同一个像素点的卷积特征不变假设下，我们结合传统的金字塔LK光流算法，构成所提出的光照鲁棒的混合光流法。在网络的训练过程中，我们提出使用深层网络辅助训练，轻量级网络实时推理的模式。一个复杂网络被用于学习图像特征可靠性图R并用于辅助轻量级网络训练。一种mask neural reprojection error （mNRE）loss 被用来提取图像的光照、视角不变的卷积特征，并且一种新的直线peak loss 被用于计算特征点得分图。
% 卷积网络适合提取图像中对光照鲁棒的描述特征。传统方法能够高效且准确的计算图像光流。二者结合即可得到同时具备高效准确并对光照鲁棒的混合光流法。

\begin{figure*}[!t]
        \centering
        \includegraphics[width=0.99\textwidth]{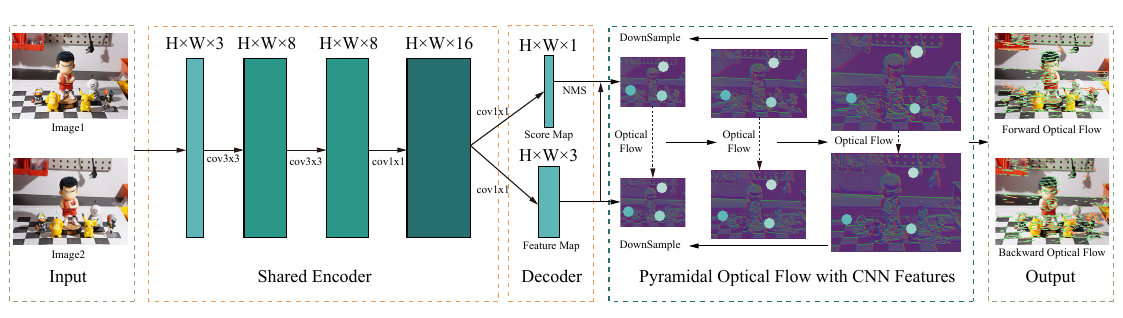}
        \caption{\textbf{The pipeline of the proposed hybrid optical flow method.} A shared encoder is first used to extract the shared feature map of the image, and then the shared feature map are decoded into score map $S$ and illumination-invariant feature map $F$. The score map $S$ is utilized for extracting keypoints, employing non-maximum suppression (NMS) to identify them. The illumination-invariant feature map $F$ are used to construct the pyramid optical flow method. Following the pyramid LK optical flow method, the hybrid optical flow method begins tracking the extracted keypoints from the highest level of the feature pyramid. The feature map are utilized to locate the positions of these keypoints in another image. Subsequently, the tracking results from the upper level are utilized as initial values for the tracking computation in the lower level, ultimately yielding sparse optical flow results.} 
        \label{fig_m1} 
\end{figure*}

We propose a hybrid optical flow method that combines deep learning with traditional approaches to overcome the limitations of conventional methods under HDR scenes. The difference between the hybrid optical flow method and LK optical flow method is clearly demonstrated in Fig. \ref{fig_i1}. Our approach begins by utilizing a lightweight convolutional neural network (CNN) to extract illumination-invariant feature maps and score maps from images. These feature maps go beyond basic RGB information, encompassing higher-level information. Subsequently, we integrate these feature maps with the traditional pyramid LK optical flow method, as Fig. \ref{fig_m1}, resulting in the hybrid optical flow method. In our approach, the training process involves two key steps. Firstly, we employ a strategy of assisted training using deep networks to enhance the performance of the lightweight network. Secondly, we introduce new loss functions such as the mask neural reprojection error (mNRE) loss to learn illumination-invariant features and the line peaky loss learn keypoint scores. The hybrid optical flow method achieves both efficient and accurate optical flow computation while demonstrating robustness to changes in illumination.
% 首先一个共享编码器用于提取图像局部特征，然后将特征解码为得分图S和卷积特征F。得分图S被用于提取跟踪角点。卷积特征F被用于构建金字塔光流，用于对角点的跟踪。
%我们的贡献主要如下：
In summary, the main contributions of this paper are as follows:

%1. 我们提出了一种基于图像学习特征不变性的光流法，打破了光照不变假设。
%2. 一种深层网络监督训练，浅层网络推理的思路，兼顾性能与效率，这使得所提出的算法在CPU下达到190Hz。
%3. 我们改进了 Peaky Loss 和 NRE 损失函数，分别用于学习易于跟踪的特征点得分图与具有光照不变性的卷积特征。

\begin{itemize}
  \item [1)] 
  We propose a hybrid optical flow method that does not rely on brightness consistency assumptions and can work properly in HDR environments. 
  \item [2)]
  We propose a loss function mNRE for extracting light invariants in images, and improve the peaky loss in keypoint extraction.
  \item [3)]
  We propose to use a deep network to train and a shallow network to infer, balancing performance and efficiency, reaching 190Hz on the CPU.
\end{itemize}

% Section \uppercase\expandafter{\romannumeral2} reviews the development of the traditional sparse optical flow method, learning-based optical flow methods, and keypoint extraction methods. Then the hybrid optical flow method is introduced in Section \uppercase\expandafter{\romannumeral3}. Section \uppercase\expandafter{\romannumeral4} presents the training process and the loss function. In Section \uppercase\expandafter{\romannumeral5}, we evaluate the performance of the proposed method in dynamic lighting conditions and integrate it into VINS-Mono. Finally, Section \uppercase\expandafter{\romannumeral6} concludes this work.

\section{RELATED WORK}
% 在本节，传统的光流法以及针对动态光照环境的改进策略的相关工作被首先介绍。改进方法通过梯度等手工计算信息提高了光流法的对光照的鲁棒性。基于学习的光流法近年来取得了更多的关注。通过端到端的方法估计图像序列的光流不依赖于任何假设，在挑战性环境下表现优秀。另一方面，所提出的方法能够同时提取关键点。因而我们同样介绍了基于学习的关键点相关工作。

% In this section, we begin by introducing traditional optical flow methods and strategies for enhancing their performance in dynamic lighting conditions. Then, the learning-based optical flow method was introduced. These methods estimate optical flow in an end-to-end manner without relying on assumptions, performing well in challenging environments. Additionally, we discuss related work on keypoint extraction methods, as our proposed approach can extract keypoints simultaneously.\par 

\subsection{Traditional optical flow method}
%Horn and Schunck提出第一个真正的光流法（HS），将光流估计定义为最小化全局能量函数的优化问题。HS使用欧拉-拉格朗日方程求解it，随后Successive Over-Relaxation（SOR）[33]和multi-grid methods[34]也被用来求解it。高斯滤波作为预处理操作被引入变分方法中，用来提高方法在噪声条件下的表现[8,35,36]。为了提高BCA的鲁棒性，[8,37]等人提出梯度一致性和其他高阶一致性约束，如the constancy of the Hessian and the constancy of the Laplacian。另外，[38, 39，40，41]发现将RGB通道变换到HIS或HSV颜色空间可以提高对光照的鲁棒性。针对显著遮挡的大位移光流计算，Revaud [10]等人提出基于匹配的方法EpicFlow。[42，43]解决了EpicFlow易受到匹配噪声影响的问题。

Horn and Schunck  (HS) \cite{hs} proposed the first truly optical flow method, which  formulated optical flow estimation as an optimisation problem minimising the global energy function. Gaussian filtering has been introduced as a pre-processing operation in variational methods to improve the performance in noisy conditions \cite{zimmer2011optic}. 
%与HS计算整张图像的光流场不同，Lucas等人[2]提出一种仅针对特定点的光流跟踪方法。LK提出邻域内空间一致性假设，并将光流估计定义为最小化邻域内加权平方和的优化问题。为了解决实际场景下像素间运动不满足小运动约束的问题，Bouguet[18]提出金字塔结构实现coarse to fine的pyramidal Lucas Kanade（PLK）。Extended Lucas Kanade（ELK）[19]认为LK算法是一个最大似然估计问题，then使用贝叶斯框架解决前后景交界处的跟踪精度问题。Robust Local Optical Flow(RLOF) [20,45]提出调整邻域窗口大小来解决广义孔径问题。作者后面又提出适用于光照变化和长距离跟踪的改进方法[21]。[44]提出选择更易于跟踪的点可以改进表现。
In contrast to HS, which computes the optical flow field for the entire image, Lucas et al. \cite{lk} proposed a method to track the optical flow for specific points. To address the issue of insufficient pixel displacement in real scenes, Bouguet \cite{bouguet} proposed a pyramid structure to implement the coarse-to-fine pyramidal Lucas Kanade (PLK). To improve the robustness of illumination, \cite{papenberg2006highly} proposed gradient consistency and other higher-order consistency constraints, such as constancy of the Hessian and constancy of the Laplacian. Robust Local Optical Flow (RLOF) \cite{robust_op, senst2016robust} proposes the adjustment of the neighbourhood window size to solve the generalised aperture problem. While employing these methods can enhance the robustness of optical flow, they are unable to fundamentally resolve the issue. \par 

\subsection{Learning-based optical flow method}
%Dosovitskiy [4]等人第一个利用卷积神经网络来计算光流。两个端到端的网络 FlowNetS和FlowNetC被提出用来从synthetic annotated dataset Flying Chairs 中直接学习光流。通过添加针对小位移的子网络等方式，[12，13]提高了FlowNet的光流估计的准确度。 [14,15]将传统方法中的金字塔模型引入网络框架，使用更少的模型参数来提高算法的运行效率。受迭代更新的启发，一种用于光流估计的新型网络架构Recurrent All-Pairs Field Transforms (RAFT)[6]被提出,并且 [46，47]在RAFT的基础上进一步提高了模型的检测效率。
CNN were first used to compute optical flow by Dosovitskiy \cite{flownet}. Two end-to-end networks, FlowNetS and FlowNetC, were proposed to learn optical flow directly from the synthetic annotated Flying Chairs dataset. The accuracy of the optical flow estimation has been improved by the addition of sub-networks for small displacements in FlowNet \cite{flownet2}. Inspired by iterative updates, Recurrent All-Pairs Field Transforms (RAFT) \cite{raft} proposed a new network architecture for optical flow estimation, and \cite{xu2022gmflow} further improved its detection efficiency. Although end-to-end methods are designed to be as lightweight as possible, the complexity of optical flow estimation still limits real-time computation to GPUs.\par 

\subsection{Keypoint detection}
%经典的Harris[22]角点检测子使用自相关矩阵来搜索角点，解决了角点各向异性和计算复杂度的问题。为了提高角点的跟踪表现，Shi and Tomasi[32]提出使角点更加分散、定位更加准确的选择标准，该方法在光流法中得到广泛应用。基于机器学习的FAST[26]检测子以牺牲角点质量为代价，极大提高了角点的检测速度。另一方面，SIFT[]、SURF[]、ORB[]、KAZE[]、AKAZE[]等具备几何不变性的角点提取方法紧接着被提出。

Classical Harris\cite{harris} keypoint detection uses the autocorrelation matrix to search for keypoints. To enhance the tracking performance of the keypoints, Shi and Tomasi \cite{good} proposed a selection criterion that makes the keypoints more distributed, which has been widely used in optical flow methods. In addition, SIFT \cite{sift}, ORB \cite{orb} etc. geometric methods were proposed to extract keypoints and matched by descriptors.\par 

%受手工制作的特征检测器的启发，基于CNN的检测的一般解决方案是构建响应图来搜索受监督的兴趣点[100，101，02]。随后SuperPoint[]建议使用预训练模型生成pseudo-ground truth points进行自我监督学习方式。进一步的，UnSuperPoint、DetNet、L2-net、D2-net、KeyNet、ECFRNet[]等无监督训练方法用来提取角点。KeyNet[30]和ALIKE[31]提出可微分的特征点检测模块用来解决NMS造成的不可反向传播优化问题。
Inspired by handcrafted feature detectors, a common approach for CNN-based detection is to construct response maps to locate interest points in a supervised manner \cite{zhang2017}. SuperPoint \cite{superpoint} subsequently suggested self-supervised learning using a pre-trained model to generate pseudo-ground truth points. Furthermore, unsupervised training methods are used to extract keypoints, KeyNet\cite{keynet} and others. To overcome the problem of non-differentiable Non-maximum suppression (NMS), ALIKE \cite{alike} propose differentiable feature point detection modules. \par 

% 所提出的方法并不属于传统光流法也不是端到端的基于学习的光流法。它将学习方法和传统方法结合，从而兼顾计算效率和性能，被我们称为混合光流法。

The proposed method does not belong to either traditional optical flow or end-to-end learning-based optical flow. It combines learning-based approaches with traditional methods, striking a balance between computational efficiency and performance. We refer to it as a hybrid optical flow method. \par 

\section{HYBRID OPTICAL FLOW METHOD}
\label{hyop}

% 0. 一段话简要介绍混合光流法的基本流程和思想。
% 1. 网络结构介绍
%  1.1 介绍Shared Encoder层
%  1.2 介绍keypoint Detector层 
%  1.3 介绍Feature Map层 
% 2. 使用卷积特征图的光流法
%  2.1 介绍光流法的基本原理
\subsection{Network architecture}

% 如图x所示，网络被设计为尽可能的轻量。它在四次卷积操作内完成了同时提取角点得分图和特征图功能。为了提高角点检测的精度以及卷积特征图的精度，我们始终在网络中保持原始分辨率。首先对输入图片提取图像特征。在网络的最后一层，我们采用了一个1x1的卷积网络将特征图的通道数降低为4。其中，特征图的前三个通道为卷积特征图，最后一个通道为角点得分图。由于网络层数少、卷积核尺寸小，因此特征图中包含的为高层语义信息较少，而是更多的保留了图像的低层信息。我们认为底层信息特征点的提取及特征图的计算仅依赖于图像的局部特征而并不需要全局信息。因此所设计的网络计算量远低于其他基于深度学习特征点。具体的每一个部分的解释如下：

As illustrated in Fig. \ref{fig_m1}, the network is designed to be as lightweight as possible which only uses four convolution operations. First, shared feature map of size $W \times H \times 16$ are extracted from the input image($W \times H \times 3$). The shared feature map is then transformed into a illumination-invariant feature map and a keypoint score map, using a $1 \times 1$ convolution kernel. With the lightweight network, the illumination-invariant feature map contains less high-level semantic information while retaining more low-level image information. We consider that the low-level information is sufficient for our tasks. Therefore, the computation complexity of the designed network is much lower than other ones. Each of these is explained in more detail as follows:

% 图像特征编码器将输入图像I转换为特征图。它仅包含三次卷积操作，每次卷积之后都采用了ReLU激活函数。前两次卷积运算采用3x3大小的卷积核，并将特征图扩展至8通道。最后一层采用1x1的卷积核将通道数量增加为16。通过感受野计算公式可以得到所提出的编码器感受野仅为5x5，与传统方法相当。我们在所有卷积操作中保持原始图像分辨率，因此最后得到WxHx16的特征图。

% 特征解码层将特征图解码为角点得分图和卷积特征图。它包含一次卷积采用，采用1x1的卷积核将特征图通道数降低为4。其中前三个通道为卷积特征图，最后一个通道为角点得分图。在卷积之后对角点得分图进行了Sigmoid激活函数，将其限制在0-1之间。对卷积特征图进行了L2归一化。最终输出结果为WxHx1的角点得分图和WxHx3的卷积特征图。

\begin{itemize}
        \item[(a)] \textbf{Shared Encoder} The image feature encoder converts the input image $I \in W \times H \times 3$ to the size $W \times H \times 16$. The first two convolution operations use a $3 \times 3$ convolution kernel and expand the shared feature map to 8 channels. In the last layer, a $1 \times 1$  convolution kernel is used to increase the number of channels up to 16. The ReLU \cite{relu} activation function is used after each convolution. We keep the original image resolution in all convolution operations.
        \item[(b)] \textbf{Feature and Socre Map Decoder} The shared feature map is decoded into a score map and an illumination-invariant feature map by the decoding layer. It uses an $1 \times 1$ convolution kernel to reduce the channels of the feature map to $4$. The first three channels are the illumination-invariant feature map, and the last channel is the keypoint score map. After the convolution, the score map is activated by the sigmoid function to limit its value between $[0,1]$. The convolution feature map is also L2 normalised. So we get the final output with a score map of $W \times H \times 1$ and a illumination-invariant feature map of $W \times H \times 3$.
\end{itemize}

% 一个浅层的网络首先被用于提取得分图S和特征图F。然后为了监督训练角点的可靠性，一个深层卷积被用于提取稠密描述子图D。最后，我们基于[S,F,D]结果计算角点损失以及匹配损失。
\begin{figure*}[!t]
        \centering
        \includegraphics[width=0.99\textwidth]{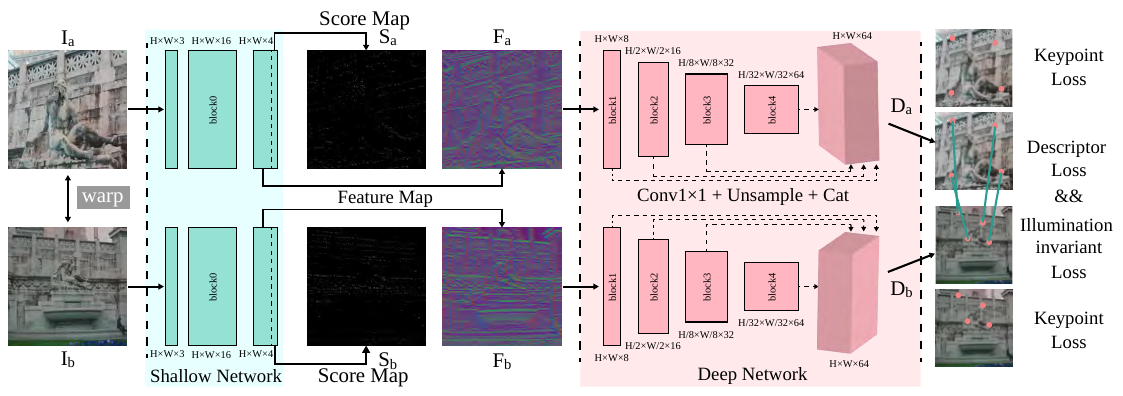}
        \caption{\textbf{The network training process.} A shallow network is first used to extract the score map $\mathbf{S}$ and feature map $\mathbf{F}$. Then, in order to supervise the reliability of the training keypoints, a deep network is used to extract the dense descriptor map $\mathbf{D}$. Finally, we calculate the keypoint loss, feature loss and descriptor loss based on the results of $[\mathbf{S},\mathbf{F},\mathbf{D}]$. Only the shallow network was used for the hybrid optical flow method depicted in Fig. 2, while the deep network was only used for training.}
        \label{fig_m2}
\end{figure*}

\subsection{Optical flow method with illumination-invariant feature maps}

% 0. 简要描述混合光流法过程
% 角点在得分图中被首先提取。我们认为得分高的位置具有强不变性并易于跟踪。在得到角点位置后，我们将其作为光流法的初始点。在LK光流的基础上，我们将亮度不变假设修改为卷积特征不变假设，即不同图像中的同一个角点位置的卷积特征向量相同。在这一假设下，结合光流金字塔算法，一个新的混合光流法被提出。

The keypoint is first extracted from the score map. Then, we use it as the initial position of the optical flow method. Based on the LK optical flow method, we modify the brightness consistency assumption to illumination-invariant feature map consistency, i.e. the convolution feature vector of the same keypoint position in different images is the same. Assuming that, a new hybrid optical flow method is proposed in combination with the pyramidal optical flow method. \par 

% 1. 角点提取

% 非极大值抑制NMS是最常用于在得分图中提取特征点的方法。然而光流法期望角点的分布更加分布均匀从而提高整体跟踪鲁棒性。因此我们采用了和OpenCV中的GoodFeaturesToTrack方法相似的方法来提取角点。首先采用非极大值抑制保留3x3邻域内的局部最大值，然后剔除小于阈值的角点。最后，为了保证角点的均匀分布，我们采用了最大间隔采样方法。具体的，我们首先将角点按照得分从大到小排序，然后从得分最高的角点开始，每隔一定的距离采样一个角点。

\subsubsection{Keypoint extraction}
The optical flow method expects a more uniform distribution of keypoints to improve the overall tracking robustness. So we use a method similar to the \textit{GoodFeaturesToTrack} function in OpenCV to extract keypoints. First, the local maximum in the 3x3 neighbourhood is retained by NMS, and then the keypoints with lower scores than the threshold are removed. We then use the maximum interval sampling method to ensure a uniform distribution of the keypoints. \par %Specifically, we start sampling a keypoint every distance from the keypoint with the highest score, after first sorting the keypoints in descending order of score. \par 

% 2. 光流计算 

\subsubsection{Pyramid optical flow method}

According to \cite{plk}, pyramid optical flow method can be divided into three steps. First, the spatial and temporal derivatives of the illumination-invariant feature map $\mathbf{F}(x,y,t)$, namely $\mathbf{F}_x$, $\mathbf{F}_y$ and $\mathbf{F}_t$, are computed. Then, the derivatives of all keypoints are combined into a coefficient matrix $\mathbf{A}$ and a constant vector $\mathbf{b}$, respectively given by
\begin{equation}
    \mathbf{A} = \left[\begin{matrix}
        \left[\begin{matrix}
            \mathbf{F}_x & \mathbf{F}_y 
         \end{matrix}\right]_1 \\
         \vdots \\
         \left[\begin{matrix}
            \mathbf{F}_x & \mathbf{F}_y 
         \end{matrix}\right]_k
     \end{matrix}\right], \mathbf{b} = \left[\begin{matrix}
        -\mathbf{F}_{t1} \\
        \vdots \\
        -\mathbf{F}_{tk}
    \end{matrix}\right],
\end{equation}
where k is the number of keypoints. Finally, the optical flow velocity $\mathbf{v}$ is obtained by solving the equation $\mathbf{A} \mathbf{v} = \mathbf{b}$. In our method, we use the standard LK optical flow algorithm, but modify the brightness constancy assumption to a convolution feature constancy assumption. Therefore, our method is called the hybrid optical flow method.

\section{LEARNING TRACKED KEYPOINTS AND INVARIANT FEATURE}
% 0. 一段话介绍如何训练角点提取网络和特征提取网络 
% 1. 学习易于跟踪的角点 
%  1.1 重投影误差 
%  1.2 line peaky loss 
%  1.3 reliability loss 
% 2. 学习图像中的不变特征
%  2.1 mask NER loss 

% 网络训练过程

% 0. 一段话介绍如何训练角点提取网络和特征提取网络
% 我们对数据集中的一组图片用相同的浅层网络参数计算得分图S，卷积特征图F，如图x所示。接着一个VGG类似的多尺度深层网络被用于提取稠密描述子图D，这被用于监督训练特征点的可靠性。基于ALike的工作，我们同样采用了重投影损失函数、重复性损失函数和peaky损失函数提取特征点得分图。此外，针对角点提取时快速收敛到直线边缘问题，我们提出了一种基于困难样本挖掘思路的line peaky loss。计算全局特征的NRE损失函数被添加了一个局部mask，得到了只考虑局部不变性的损失函数。
Fig. \ref{fig_m2} illustrates the training process of the network. For an image pair, a shallow network is used for the extraction of the score map $\mathbf{S}$ and the feature map $\mathbf{F}$, and a deep network is used for the extraction of the dense descriptor map $\mathbf{D}$. The shallow network here is consistent with the network structure in Sec. \ref{hyop} while the deep network is only used to assist in training. The keypoint loss, the illumination invariant feature loss, and the descriptor loss are used to train the three distinct outputs $[\mathbf{S},\mathbf{F},\mathbf{D}]$, respectively. The keypoint loss consists of reprojection loss, line peaky loss and reliability loss. The NRE function and the proposed mNRE function are used for descriptor loss and illumination invariant feature loss, respectively. \par

% 学习具有强不变性的角点
\subsection{Keypoint loss}
% 三个损失函数被用来训练角点的提取。重投影损失函数和peaky损失函数用来保证角点的准确性，并利用稠密描述子图D来监督角点的可靠性。具体的损失函数如下所示：
As described in previous work\cite{alike}, a good keypoint should be repeatable, highly accurate, and matchable. Thus, three loss functions are used to train for the extraction of keypoints. The reprojection loss function ensures the repeatability of the keypoints. The line peaky loss function is beneficial for accuracy improvement of keypoints. And loss of reliability makes it easier to match the extracted keypoints.
\subsubsection{Reprojection loss}
% 1. 重投影误差
% 图像$\mathbf{I}_A$中的一点$\mathbf{p}_a$被投影至图像$\mathbf{I}_B$中，投影点为$\mathbf{p}_{AB}$。假设$\mathbf{p}_a$和$\mathbf{p}_{AB}$之间的距离为$\mathrm{dist}_{AB}$，则重投影误差为：
% 一个关键点应该在不同图像中具有相同的位置。
% 
A keypoint should be extracted simultaneously in two images under different conditions. The reprojection error is defined as the distance between the projected point and the extracted point. A point $\mathbf{p}_A$ in image $\mathbf{I}_A$ is projected to image $\mathbf{I}_B$, and the projection point is $\mathbf{p}_{AB}$. The single reprojection error is
\begin{equation}
        \mathrm{dist}_{AB} = \|\mathbf{p}_{AB} - \mathbf{p}_B\|,
\end{equation}
where $\mathbf{p}_B$ is the extracted point in image $\mathbf{I}_B$ and $\| \cdot \|$ is 2-norm of the vector. 
% 重投影误差损失函数被定义对称的形式：
The reprojection error loss is defined in a symmetrical form
\begin{equation}
        \mathcal{L}_{rp} = \frac{\sum_{i=0}^{N-1}\mathrm{dist}^{i}_{AB}}{N} + \frac{\sum_{i=0}^{M-1} \mathrm{dist}^{i}_{BA}}{M},
\end{equation}
% 其中，N是图像A中提取的关键点中能够在图像B中找到的点的数量，M是图像B中提取的关键点中能够在图像A中找到的点的数量。
where $N$ is the number of points in image $\textbf{I}_A$ that can be found in image $\textbf{I}_B$, $\mathrm{dist}^{i}_{AB}$ is the reprojection error of the ith feature out of the $N$ features,  $M$ is the number of points in image $\textbf{I}_B$ that can be found in image $\textbf{I}_A$, and $\mathrm{dist}^{i}_{BA}$ is the reprojection error of the ith feature out of the $M$ features.
% 2. line peaky loss
% 四种直线模型分别代表垂直、水平、左斜、右斜四种直线。图中展示的权重数值为距离d(i,j)与公式x的乘积。

% \begin{figure}[!t]
%         \centering
%         \includegraphics[width=0.48\textwidth]{img/3.pdf}
%         \caption{\textbf{Visualisation of the weights of the four line models.} Four line models represent vertical, horizontal, left oblique, and right oblique lines, respectively. The weight values shown in the figure are the product of the distance $d(i,j)$ and the formula \ref{eq_m1}.} 
%         \label{fig_m3} 
%         \end{figure} 

\subsubsection{Line peaky loss}
\begin{figure}[!t]
        \centering
        \includegraphics[width=0.48\textwidth]{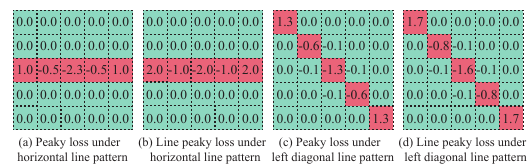}
        \caption{\textbf{Comparison of line peaky loss and peaky loss.} In $5 \times 5$ sized patch, the cyan block represents score 0.5 while the red represents score 1. The numbers in the blocks represent then the derivative of the different losses with respect to the block. It can be seen that the line peaky loss increases the penalty weight for the ends of the lines.} 
        \label{fig_line} 
        \end{figure} 
On the score map, keypoints should take on a peaked shape in the neighborhood. Consider a $N \times N$-sized patch near the keypoint $\textbf{p}$ on the score map, the distance between each pixel location $[i, j]$ and the keypoint is
\begin{equation}
    d(\textbf{p}, i, j)=\left\{\|\textbf{p} - [i, j] \| \right\},
\end{equation}
where $\| \cdot \|$ is a 2-norm of the vector. The peaky loss proposed by \cite{alike} is used to reduce the score of farther-distant positions within a patch, defined as
\begin{equation}
    \mathcal{L}_{pk}(\textbf{p})=\frac{1}{N^2} \sum_{0\leq i, j, < N } d(\textbf{p}, i, j) s(\textbf{p}, i, j),
\end{equation} 
where $s(\textbf{p},i,j)$ is the score corresponding to the position $[i, j]$ in the patch near the keypoint $\textbf{p}$. This definition views all pixels within the patch uniformly, making the score map form a locally linear shape during training. Thus, we consider four line patterns, horizontal, vertical, left diagonal, and right diagonal, with increased penalty weights for line shapes. In a patch of size $N \times N$ near the keypoint $\textbf{p}$, the four line weights $w_1,w_2,w_3,w_4$ are defined as
\begin{equation}
        \begin{aligned}
                w_1(\textbf{p},i,j) &= \mathcal{N}(|i - p_x|)\\
                w_2(\textbf{p},i,j) &= \mathcal{N}(|j - p_y|)\\
                w_3(\textbf{p},i,j) &= \mathcal{N}(|i+j - p_x - p_y|)\\
                w_4(\textbf{p},i,j) &= \mathcal{N}(|i-j - p_x + p_y|)
        \end{aligned},
        \label{eq_m1}
\end{equation}
where $[i, j]$ is the pixel position within the patch, $p_x, p_y$ are the coordinates of the keypoint $\textbf{p}$, $\mathcal{N}$ is the gaussian distribution and $|\cdot|$ is the 1-norm. By using these weights, the line peaky loss are defined as
\begin{equation}
\begin{aligned}
    &\mathrm{s}(\mathbf{p}, i, j) = w_k(\mathbf{p}, i, j) d(\mathbf{p}, i, j) s(\mathbf{p}, i, j) \\
    &\mathcal{L}_{lpk}(\mathbf{p}) = \max_{k=1 \cdots 4} \left\{ \frac{1}{N^2} \sum_{[i, j]^T \in patch(\mathbf{p})} \mathrm{s}(\mathbf{p}, i, j) \right\}
\end{aligned}, 
\end{equation}
where the $\max$ function is used to select the maximum value from the four line patterns. Fig. \ref{fig_line} demonstrates that the peaky loss differs from the derivative of the proposed line peaky loss in that the penalty at the ends of the line is increased by the line peaky loss.
\subsubsection{Reliability loss}
% 准确且可重复的关键点是不够的，还需要保证关键点的可匹配能力，即可靠性。因此，我们采用了稠密描述子图D的匹配结果得到角点的可靠性。具体的，我们首先计算角点的可靠性$r_{\mathbf{p}_A}$：

Accurate and repeatable keypoints are not sufficient. It is also necessary to ensure the matchability of the keypoints \cite{ alike}. To compute the matchability of the keypoint $\textbf{p}_{A}$ in image $\textbf{I}_A$, the vector distance between its corresponding descriptor $\textbf{d}_{\textbf{p}_A} \in \mathbb{R}^{dim}$ and the dense descriptor map $\textbf{D}_B \in \mathbb{R }^{H \times W \times dim}$ in image $\textbf{I}_B$ are computed as
\begin{equation}
    \label{equ_c}
    \mathbf{C}_{\textbf{d}_{\textbf{p}_A}, \textbf{D}_B} = \textbf{D}_B  \textbf{d}_{\textbf{p}_A} ,
\end{equation}
where $\mathbf{C}_{\textbf{d}_{\textbf{p}_A}, \textbf{D}_B} \in \mathbb{R}^{H \times W}$ is known as the similarity map representing the similarity between the keypoint $\textbf{p}_{A}$ and the position of each pixel in the image $\textbf{I}_B$. The normalization function then makes the score of the positions with high similarity to $1$, while the score of the positions with low similarity is $0$, thus the matching probability map $\widetilde{\mathbf{C}}_{\textbf{d}_{\textbf{p}_A}, \textbf{D}_B} \in \mathbb{R}^{H \times W}$ is defined as
\begin{equation}
    \widetilde{\mathbf{C}}_{\textbf{d}_{\textbf{p}_A}, \textbf{D}_B} = \exp \left(\frac{\mathbf{C}_{\textbf{d}_{\textbf{p}_A}, \textbf{D}_B}-1}{t} \right) ,
\end{equation}
where $\exp$ is the exponential function used to compose the normalization function, and $t=0.02$ controls the shape of the function. For the keypoint $\textbf{p}_{A}$, the higher matching probability of its projected location $\mathbf{p}_{AB}$ implies its higher reliability. Therefore the reliability $r_{\mathbf{p}_A}$ of the keypoint $\mathbf{p}_A$ is defined as
\begin{equation}
\label{eq_sample}
    r_{\mathbf{p}_A}=\mathbf{bisampling} \left(\widetilde{\mathbf{C}}_{\textbf{d}_{\textbf{p}_A}, \textbf{D}_B}, \mathbf{p}_{AB}\right) ,
\end{equation}
where $\mathbf{bisampling}(\mathbf{M}, \mathbf{p})$ is a function of bilinear sampling of the $\mathbf{M} \in \mathbb{R}^{H \times W}$ at position $\mathbf{p}\in \mathbb{R}^{2}$. Then consider all the keypoints in the image $\textbf{I}_A$ and penalize the keypoint scores that are less reliable among them to get the reliability loss
\begin{equation}
    \begin{aligned}
        \mathbb{S} &= \sum_{\mathbf{p}_A \in \mathbf{I}_A \atop \mathbf{p}_{AB} \in \mathbf{I}_B } s_{\mathbf{p}_A} s_{\mathbf{p}_{AB}} \\
        \mathcal{L}_{reliability}^{A} &= \frac{1}{N_A} \sum_{\mathbf{p}_A \in \mathbf{I}_A \atop \mathbf{p}_{AB} \in \mathbf{I}_B } \frac{s_{\mathbf{p}_A} s_{\mathbf{p}_{AB}}}{\mathbb{S}}(1-r_{\mathbf{p}_A})
    \end{aligned} ,
\end{equation}
where $N_A$ is the number of all keypoints in the image $\mathbf{I}_A$, $s_{\mathbf{p}_A}$ is the score of the keypoint $\mathbf{p}_A$ and $s_{\mathbf{p}_{AB}}$ is the score of the projection location.

Based on the above three loss functions, we obtain the keypoint loss
\begin{equation}
    \begin{aligned}
        \mathcal{L}_{keypoint} = &k1 \cdot \mathcal{L}_{rp} + \\ 
        &k2 \cdot\frac{1}{N+M} (\sum_{\mathbf{p}_A \in \mathbf{I}_A} \mathcal{L}_{lpk}(\mathbf{p}_A)+ \sum_{\mathbf{p}_B \in \mathbf{I}_B } \mathcal{L}_{lpk}(\mathbf{p}_B)) + \\ 
        & k3 \cdot \frac{1}{2}(\mathcal{L}_{reliability}^{A}+\mathcal{L}_{reliability}^{B}) 
    \end{aligned} ,
\end{equation}
where $k_1=1,k_2=0.5,k_3=1$ are the weights and $N,M$ are the number of keypoints in image $\mathbf{I}_A$ and $\mathbf{I}_B$ respectively.

\subsection{Descriptor loss}
The NRE \cite{nre} function was used to learn the descriptors. Due to its good performance, we adopted it as the descriptor loss. Previous work \cite{alike} explains the derivation and definition of NRE function by the cross-entropy function. We give an explanation from another more intuitive perspective. Based on the similarity map defined in Eq. \ref{equ_c}, a new matching probability map $\widetilde{\mathbb{C}}_{\textbf{d}_{\textbf{p}_A}, \textbf{D}_B}\in \mathbb{R}^{H \times W}$ is defined as
\begin{equation}
\label{eq_match}
    \widetilde{\mathbb{C}}_{\textbf{d}_{\textbf{p}_A}, \textbf{D}_B}:= \mathbf{softmax} \left( \frac{C_{\textbf{d}_{\textbf{p}_A}, \textbf{D}_B}-1}{t} \right)  ,
\end{equation}
where the normalization function $\mathbf{softmax}$ converts similarity to probability and satisfies that all elements sum to one, i.e., $\sum_{H \times W} \widetilde{\mathbb{C}}_{\textbf{d}_{\textbf{p}_A}, \textbf{D}_B} = 1$. For a good descriptor $\mathbf{d}_{\mathbf{p}_A}$, it should be as similar as possible to the descriptor at projection position $\mathbf{p}_{AB}$ and far away from all other descriptors. Thus, by the same sampling function in Eq. \ref{eq_sample}, we get the matching probability $\mathrm{p}^{\mathbf{p}_{AB}}_{\textbf{d}_{\textbf{p}_A}, \textbf{D}_B}$ as
\begin{equation}
    \mathrm{p}^{\mathbf{p}_{AB}}_{\textbf{d}_{\textbf{p}_A}, \textbf{D}_B} = \mathbf{bisampling} \left(\widetilde{\mathbf{\mathbb{C}}}_{\textbf{d}_{\textbf{p}_A}, \textbf{D}_B}, \mathbf{p}_{AB}\right).
\end{equation}
Maximizing the matching probability at the projected positions $\mathrm{p}^{\mathbf{p}_{AB}}_{\textbf{d}_{\textbf{p}_A}, \textbf{D}_B}$ with the constraint that the sum of all the elements is equal to $1$ implies minimizing the matching probability at the other positions. The descriptor loss function is then obtained as
\begin{equation}
    \mathcal{L}_{desc}=\frac{1}{N_A+N_B}\cdot (-\sum_{\mathbf{p}_A\in \mathbf{I}_A} \ln (\mathrm{p}^{\mathbf{p}_{AB}}_{\textbf{d}_{\textbf{p}_A}, \textbf{D}_B}) -\sum_{\mathbf{p}_B\in \mathbf{I}_B} \ln (\mathrm{p}^{\mathbf{p}_{BA}}_{\textbf{d}_{\textbf{p}_B}, \textbf{D}_A})) ,
\end{equation}
where $N_A$ is the number of keypoints in image $\mathbf{I}_A$, $N_B$ is the number of keypoints in image $\mathbf{I}_B$, and the $-\ln(\cdot)$ function converts the maximization problem into a minimization problem. 

\subsection{Illumination-invariant feature loss}
The illumination invariant feature map is similar to the dense descriptor map in that both attempt to extract features in an image that do not vary with viewing point and illumination. The difference is that descriptors are usually obtained using 64 or 128 channels and are convolved several times, whereas illumination invariant feature maps have only three or one channel and are obtained by four convolutions. Thus illumination invariant feature maps are difficult to learn to obtain features that are distinguishable over the entire image. Therefore we propose to learn locally distinguishable illumination invariant feature maps using mNRE loss function. Consider first defining the mask function near the keypoint $\mathbf{p}=[p_x, p_y]^T$
\begin{equation}
    \mathrm{mask}(\mathbf{p}) = \begin{cases}
                1 & \text{if } \max (|x - p_x|,|y - p_y|) < d \\
                0 & \text{otherwise}
                \end{cases},
\end{equation}
where $d=80$ is the mask range and $|\cdot|$ is the 1-norm. This mask is then added to Eq. \ref{eq_match} of the NRE loss function to obtain the local matching probability map
\begin{equation}
    m\widetilde{\mathbb{C}}_{\textbf{d}_{\textbf{p}_A}, \textbf{D}_B}:= \mathbf{softmax} \left( \frac{\mathrm{mask}(\mathbf{p}_A) \cdot C_{\textbf{d}_{\textbf{p}_A}, \textbf{D}_B}-1}{t} \right) ,
\end{equation}
where $m\widetilde{\mathbb{C}}_{\textbf{d}_{\textbf{p}_A}, \textbf{D}_B} \in \mathbb{R}^{H \times W}$ has a value of zero in the region away from $\mathrm{p}_A$. Then similar to in the NRE loss, we computed the local matching probability of the projected positions
\begin{equation}
    m\mathbb{p}^{\mathbf{p}_{AB}}_{\textbf{d}_{\textbf{p}_A}, \textbf{D}_B} = \mathbf{bisampling} \left(m\widetilde{\mathbf{\mathbb{C}}}_{\textbf{d}_{\textbf{p}_A}, \textbf{D}_B}, \mathbf{p}_{AB}\right).
\end{equation}
Finally, the loss function is computed based on the local matching probability of all keypoints in images $\mathbf{I}_A$ and $\mathbf{I}_B$ as
\begin{equation}
\begin{aligned}
    &\mathcal{L}_{feat}= \\ &\frac{1}{N_A+N_B}\cdot (-\sum_{\mathbf{p}_A\in \mathbf{I}_A} \ln (m\mathrm{p}^{\mathbf{p}_{AB}}_{\textbf{d}_{\textbf{p}_A}, \textbf{D}_B}) -\sum_{\mathbf{p}_B\in \mathbf{I}_B} \ln (m\mathrm{p}^{\mathbf{p}_{BA}}_{\textbf{d}_{\textbf{p}_B}, \textbf{D}_A})).
\end{aligned}
\end{equation}

\section{EXPERIMENTS}
% 实验 
% 大纲
% 0. 数据集介绍、训练介绍
% 1. 重复率实验 （一个对比表格）
% 2. 静态图像匹配（一个图像展示图，一个表格），动态光照匹配 （一个图片包含曲线和图片展示图）
% 3. 动态光照vins实验 （一个表格+一个带曲线+图片的展示图）

% 实验部分概述

% 在本节中，我们首先介绍了采用的数据集以及训练的细节。为了验证所提出的方法，我们首先在HPatches数据集上对比了角点的重复率。然后多组动态光照环境下采集的图像被用来验证其在HDR环境下的性能。最后，所提出的光流被嵌入至先进的VIO系统中，并对比轨迹精度，验证了其在实际应用中的有效性。

% In this section, we first introduce the training details. To validate the proposed method, we first compare the keypoint repeatability on the HPatches \cite{hpatches} dataset. Then, several image sets captured under dynamic lighting environments are used to verify the proposed optical flow method performance in HDR environments. Then the proposed method is embedded in a VIO system and the trajectory accuracy is compared to verify its effectiveness in practical use. Finally, ablation experiments verify the validity of the proposed mNRE loss and line peaky loss.

% 图像序列中的光流对比。在图像序列中，一个主动光源被用于模拟动态光照环境。在椭圆形光斑的边缘，亮度恒定假设不再成立，因此光流的性能将受到挑战。当前关键点被绘制为绿色，十帧内的光流被绘制为红色。序列中的图像每隔十帧被绘制出来。可以看到，所传统的LK光流在光斑边缘出现了较大的误差。而所提出的光流在光斑边缘的性能得到了显著的提升。

\begin{figure*}[!t]
        \centering
        \includegraphics[width=0.99\textwidth]{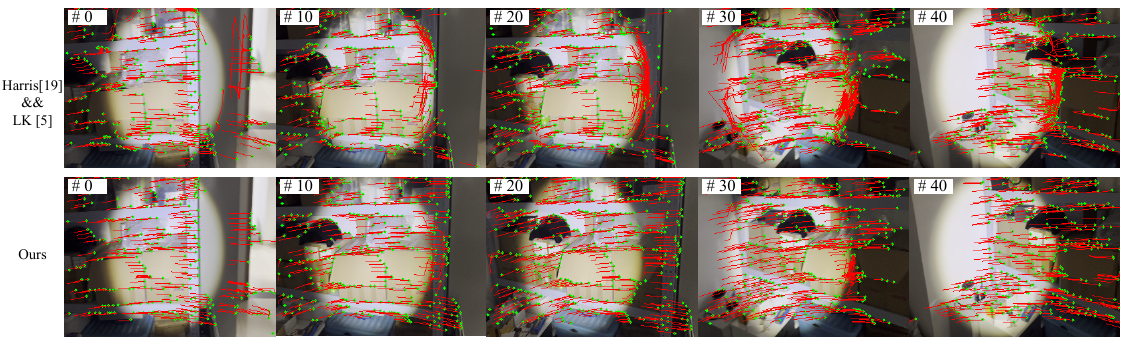}
        \caption{\textbf{Optical flow method comparison in image sequences.} In the image sequence, an active light source is used to simulate dynamic lighting environments. At the edge of the elliptical spot, the constant brightness assumption no longer holds, so the performance of the optical flow method will be challenged. The current keypoints are drawn in green, and the optical flow results within ten frames is drawn in red. Image pairs in the sequence are drawn every ten frames. It can be seen that the traditional LK optical flow method has a large error at the edge of the spot. The proposed optical flow method has significantly improved performance at the edge of the spot.} 
        \label{fig_s8} 
\end{figure*}  

% 0. 数据集介绍
%\subsection{Dataset}

% MegaDepth 数据集由135个著名景点的图像组成，共包含63k图像。通过三维重建算法COLMAP构建场景的三维地图，从而得到图像之间的位姿关系、深度信息以及密集对应关系。我们采用DISK中生成的图片来训练我们的模型。

%\textbf{MegaDepth} dataset consists of 63k images from 135 famous scenes. TThe 3D map of the scene is generated by the 3D reconstruction algorithm COLMAP to obtain the corresponding poses, depths and densities between frames. We use the images generated as DISK to train our model.

%\par 

% HPatches数据集是一个用于评估图像匹配性能的数据集。它包含57个光照场景和59个视角场景的平面图像。其中，每个场景包含5对图像，并提供了真实的H矩阵。我们采用HPatches数据集来验证所提出的角点的重复率。

%\textbf{HPatches} dataset is a dataset for evaluating image matching performance. It contains planar images of 57 illumination scenes and 59 viewpoint scenes. Each scene contains 5 pairs of images and provides the real homography matrix. We use the HPatches dataset to verify the repeatability of the proposed keypoint.

%\par 

% UMA-VI 数据集是一个用于评估动态光照及弱纹理下VIO性能的数据集。它包含了33个不同场景下的双目图像序列以及IMU数据。所有传感器的外参和内参均得到了仔细的标定。其中包含一组典型的室内-室外穿越场景，共5个光照变化序列。该序列被用来验证所提出的光流在典型光照变化下的性能。

%\textbf{UMA-VI} dataset is a dataset to evaluate the performance of VIO under dynamic lighting and weak texture. It contains image sequences and IMU data in 33 different sequences. The extrinsic and intrinsic parameters of all sensors are carefully calibrated. It contains a typical indoor/outdoor crossing scene, consisting of 5 sequences with changing lighting. This sequence is used to verify the performance of the proposed optical flow under typical lighting changes.

% 0. 训练介绍 
\subsection{Training Details}

% 训练过程中所有数据集的图像被缩放至480x480的大小。使用ADAM优化器进行的训练,学习率设置为0.01。我们设置Batch Size为1，但是采用累计16个batches的梯度，训练了100个epoch。在如上设置下，我们的模型在一张4090显卡下训练了大约1天。

MegaDepth\cite{megadepth} is used to train the model. All images in the dataset are scaled to $480 \times 480$ during the training process. Training is done using the ADAM \cite{adam} optimiser with a learning rate of $3e^{-3}$. We set the batch size to one, but use the gradient of 16 batches to accumulate, and train for 100 epochs. Using the above settings, our model was trained on a 4090 graphics card for approximately one day.
 
\begin{table}[t]
        \begin{center}
                \caption{repeatability comparison}
                \label{te1}
                \setlength{\tabcolsep}{1mm}{
                \begin{tabular}{cccc}
                        \toprule[1pt]
                        % \midrule[0.5pt]
                        \multirow{2}{*}{Method}  &  \multicolumn{2}{c}{Repeatability $\uparrow$} & \multirow{2}{*}{CPU time(ms) $\downarrow$} \\
                        \cmidrule{2-3} 
                        ~& Illumination Scenes &  Viewpoint Scenes & ~\\ 
                        \midrule
                        Ours & 0.618 & \textbf{0.606} & 5.2 \\ 
                        ALike(T) \cite{alike} & \textbf{0.638} & 0.563 & 84.4 \\
                        SuperPoint \cite{superpoint} & 0.652 & 0.503 & 93.5 \\
                        Harris \cite{harris} & 0.62 & 0.556 & 4.9 \\
                        Fast \cite{fast} & 0.575 & 0.552 & 0.4 \\
                        Random & 0.101 & 0.1 & / \\
                        \bottomrule[1pt]
                \end{tabular}}
        \end{center}
\end{table} 
% 1. 重复率实验 （一个对比表格）
\subsection{HPatches Repeatability}

% 为了验证所提出的模型检测角点的性能，我们在HPatches数据集上计算了角点的重复率。我们对比了所提出的方法与ALike(T)、SuperPoint、Harris、Fast、Random等方法。其中ALike(T)和SuperPoint是基于深度学习的特征点检测和描述子提取方法。重复率在240x320分辨率下提取300个角点时被计算。为了抑制特征点的扎堆现象，我们对所有的特征点检测方法使用了相同的非极大值抑制。两张图片中重投影距离小于3的特征点被认为是重复角点，而超过这一阈值的角点则是不可重复提取的。表格中展示了在光照场景和视角场景下的重复率对比。从表格中可以看出，所提出的方法在光照场景下的重复率达到了0.638，超过了其他方法。在视角场景下，所提出的方法的重复率为0.606，仅次于Harris方法。

In order to verify the performance of the proposed model in the detection of keypoints, we calculated the repeatability of the keypoints on the HPatches \cite{hpatches} dataset. We compared the proposed method with ALIKE(T)\cite{alike}, SuperPoint\cite{superpoint}, Harris \cite{harris}, Fast\cite{fast}, Random. The repeatability is calculated for the extraction of 300 keypoints at a resolution of $240\times320$. The same NMS is used for all feature point detection methods to suppress the phenomenon of feature point clustering. The keypoints with reprojection distance less than $3$ in the two images are considered to be repeat keypoints and the keypoints above this threshold are not repeatable. The comparison of repeatability in illumination scenes and viewpoint scenes is shown in the table \ref{te1}. We can see that the proposed keypoint point repetition rate is at the state-of-the-art level. \par

% 在计算时间上，所有的方法均在相同的CPU上运行并测量得到。我们在图像尺寸480x640计算了非深度学习的角点提取方法，而在240x320的分辨率下计算了深度学习的角点得分图，并上采样至480x640。从表格中可以看出，所提出的方法相较于其他深度学习方案实时性得到的巨大的提升，甚至与传统方法相当。

In terms of calculation time, all methods are run on the same onboard CPU I7-1165G7 and measured. We computed the non-deep learning keypoint extraction method in the size of $480 \times 640$, and computed the deep learning keypoint score map in $240 \times 320$, and then upsampled to $480 \times 640$. From the table \ref{te1}, it can be seen that the proposed method has greatly improved the real-time performance compared to other deep learning solutions, and even comparable to traditional methods.

% 2. 角点跟踪实验 （一个图像展示图，一个表格） （一个图片包含曲线和图片展示图）
\subsection{Keypoint Tracking}
\label{kp_tr}

% 采集的数据包含四个典型场景，室内光源变化、室外阳光变化、主动光源以及光照散射导致的图像模糊。在所有的图像对中均不满足灰度的假设。
\begin{figure}[!t]
        \centering
        \includegraphics[width=0.48\textwidth]{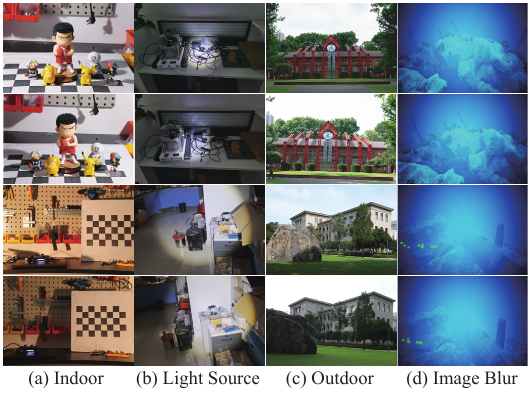}
        \caption{\textbf{Examples of dynamic lighting scene images.} The collected data contains four typical scenes, indoor light source changes, active light sources, outdoor sunlight changes and image blur caused by light scattering. The gray scale assumption is not satisfied in all image pairs.} 
        \label{fig_e1} 
        \end{figure}   
\begin{table}[htp]
        \begin{center}
                \caption{correct tracking ratio}
                \label{te2}
                \setlength{\tabcolsep}{1mm}{
                \begin{tabular}{ccccc}
                        \toprule[1pt]
                        Scenes & Indoor & Lighting Source & Outdoor & Image Blur  \\ 
                        \midrule
                        Ours & \textbf{0.84} & \textbf{0.87} & \textbf{0.67} & \textbf{0.75} \\ 
                        LK Optical Flow \cite{plk} & 0.58 & 0.37 & 0.45 & 0.19 \\ 
                        Census & 0.53 & 0.72 & 0.61 & 0.41 \\ 
                        Histogram equalization & 0.51 & 0.50 & 0.52 & 0.49 \\ 
                        ORB \cite{orb} & 0.12 & 0.40 & 0.23 & 0.32 \\ 
                        ALIKE(T) \cite{alike} & 0.59 & 0.41 & 0.54 & 0.51 \\ 
                        SuperPoint \cite{superpoint} & 0.48 & 0.49 & 0.57 & 0.51 \\ 
                        \bottomrule[1pt]
                \end{tabular}}
        \end{center}
\end{table}

% 为了验证所提出的光流法在动态光照场景下的性能，我们采集了多组具有典型的光照变化图像对以及带有主动探照光源的图像序列。图x所展示的是采集部分图像对示例，它们的共同特点为均不满足灰度不变性假设。我们将这些场景分为典型的四类，即室内光源变化、室外阳光变化、主动光源以及光照散射导致的图像模糊。在每一类场景中采集五组图像对，每组图像对中包含两张角度差距较小但是光照变化的图像。

We collected several sets of image pairs with typical lighting changes and active light sequences. Fig. \ref{fig_e1} shows some examples of collected images. These scenes are divided into four typical categories, namely indoor light source changes, outdoor sunlight changes, active light sources and image blur. Each category collects five pairs of images with small angle differences but large illumination changes.

% 基于所采集的数据，我们对比了所提出的方法与LK光流法、ORB、ALIKE(T)、SuperPoint等方法的匹配结果。所有的方法运行在480x640大小的图片上，并限制提取最多300个角点进行匹配。ORB等所有基于描述子的匹配算法均采用暴力匹配方法得到匹配结果。由于真实值的缺失，我们基于SIFT特征点对每组图像进行匹配并基于RANSAC算法估计基础矩阵。在多次剔除离群点匹配结果后，得到图像对的基础矩阵。利用该矩阵对所有的匹配结果进行筛选，并得到正确跟踪率：

Using the data, we compared the proposed methods' matching results in Table \ref{te2}. All methods run on images sized $480\times640$ and restrict to extracting up to 300 keypoints for matching.
Brute force matching is used to obtain matching results for all keypoints with descriptors such as ORB \cite{orb}. The optical flow method after preprocessing the image using census transform and histogram equalization was similarly compared. As there is no ground truth, we match each set of frames using SIFT\cite{sift} keypoints and estimate the fundamental matrix in RANSAC \cite{ransac}. This matrix serves to filter all matches to obtain the correct number.
% 表x中展示了在不同场景下的正确跟踪率对比。从表格中可以看出，所提出的方法在所有的场景下均取得了最高的正确跟踪率。我们认为这是两方面的原因导致的。首先，光流法本身具有较强的匹配能力，尤其在视角变化较小的场景下。其次，所提出的算法提升了其对光照的鲁棒性，从而取得最优的匹配结果。从表x中可以看出，原始的光流法的匹配性能在不同数据集之间变化较大，这是由于传统光流法对光照变化的敏感性导致的。
Table \ref{te2} shows the comparison of the correct tracking rate in different scenes. From the table, we can see that the proposed method has achieved the highest correct tracking rate in all scenes.

% 我们同样采集了一个带有主动光源的序列，并持续的跟踪序列中提取的特征点位置，如在VIO中一样。图x展示出我们所提出的方法可以在光照变化的场景下对特征点持续的跟踪。而传统的光流法在主动光源的干扰下，角点跟踪快速的失败导致拒绝率快速上升。

\begin{figure}[!t]
        \centering
        \includegraphics[width=0.48\textwidth]{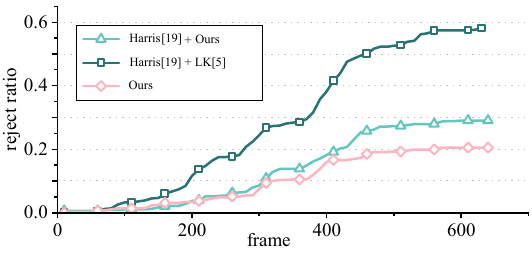}
        \caption{\textbf{Keypoint tracking rejection rate.} In the sequence with active light source, the number of outliers in the optical flow matching is counted and divided by the total number of keypoints to obtain the rejection rate. Compared with the original LK optical flow method, the proposed method can effectively reduce the number of outliers, thereby improving the accuracy of optical flow.} 
        \label{fig_e2} 
        \end{figure}

For further verification of the performance of the proposed method in dynamic lighting scenes, we also collected a sequence with an active light source. As in the VIO system, we continuously tracked the positions of the extracted keypoints in the sequence. As shown in Fig. \ref{fig_s8}, the proposed method can continuously track keypoints at the edge of the spot. Furthermore, Fig. \ref{fig_e2} shows that the proposed method can continuously track keypoints in scenes with illumination changes. The traditional optical flow method fails due to the interference of the active light source, resulting in a rapid increase in the rejection rate.

% 角点跟踪拒绝率。在含有主动光源的序列中统计光流匹配中的离群点数量并处以总的特征点数量，得到拒绝率。所提出的方法相较于原始的LK光流法，能够有效的减少离群点的数量，从而提高了光流的准确性。并且在下面以图片的形式展示了二者的光流跟踪性能差异。

% 3. 动态光照下的VIO轨迹估计 （一个表格+一个带曲线+图片的展示图）
\subsection{VIO Trajectory Estimation}

% 我们将所提出的光流法嵌入至现代VIO系统，VINS-Mono中。通过替换原始的角点提取算法以及光流计算方法得到了修改的VIO系统。在UMA-VI中包含动态光照数据集上进行了对比测试，如图图x所示。其中ORB-SLAM3表现最差，一次又一次的跟踪失败。VINS-Mono在光照变化的情况下，轨迹误差快速累计，而所提出的方法则展示出最高的精度。由于数据集只提供了轨迹初始位置和结束位置部分的真实值，我们只计算了末端结束轨迹的累计平移误差，如表x所示。

\begin{figure}[!t]
        \centering
        \includegraphics[width=0.48\textwidth]{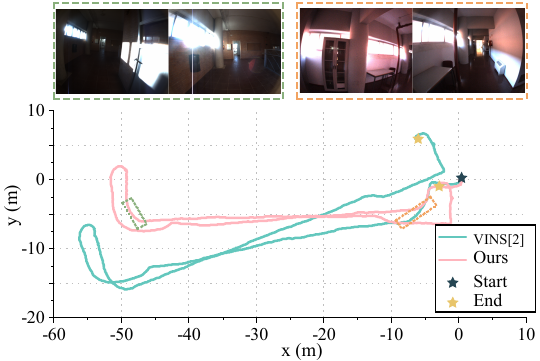}
        \caption{\textbf{Sequence 1 trajectory comparison in the UMA-VI \cite{uma-vi} dataset.} The starting point and end point of the sequence coincide, and it can be seen that the improved method can effectively improve the accuracy of the trajectory in the dynamic lighting scene. The two set of pictures corresponding to the two parts of the rapid accumulation of trajectory error are displayed above the curve, which verifies that the rapid change of lighting is the main source of error.} 
        \label{fig_e4} 
        \end{figure}

We embed the proposed optical flow method into the modern VIO system, VINS-Mono \cite{vins-mono}. By replacing the original keypoint extraction algorithm and the optical flow method calculation method, the modified VIO system is obtained. As shown in Fig. \ref{fig_e4}, a comparison test was performed on the UMA-VI \cite{uma-vi} dataset with dynamic lighting data. As the dataset only provides the part of the ground truth, we only compute the final trajectory translation error, as shown in table \ref{te3}. We similarly tested this in the widely used EuRoC \cite{euroc} dataset, as in table \ref{te5}. The results show that hybrid optical flow can equally improve the accuracy even in common datasets. \par

% UMA-VI 序列1的轨迹图。序列的起点和终点重合，可以看出改进的方法能够有效的提高轨迹在动态光照场景下的精度。并且在曲线的上方展示了轨迹误差快速累计的两个部分对应的图片，验证了光照的快速变化是误差的主要来源。

% 另一个场景的轨迹图，同样展示了光照的剧烈变化对导致轨迹误差快速累计。我们的方法不依赖于光照不变的假设，稳定跟踪角点，从而提高轨迹的精度。        
% \begin{figure}[!t]
%        \centering
%        \includegraphics[width=0.48\textwidth]{img/7-1.pdf}
%        \caption{\textbf{Sequence 2 trajectory comparison in the UMA-VI \cite{uma-vi} dataset.} The starting and ending points of the sequence coincide, so the trajectory error can be visually observed. The two set of pictures corresponding to the two parts of the rapid accumulation of trajectory error are displayed above the curve, which demonstrates that the rapid change of lighting is the main source of error.} 
%        \label{fig_e3} 
%        \end{figure}    

\begin{table}[!h]
        \begin{center}
                \caption{trajectory error comparison in indoor-outdoor dynamic illumination category}
                \label{te3}
                \setlength{\tabcolsep}{2.5mm}{
                \begin{tabular}{cccc}
                        \toprule[1pt]
                        % \midrule[0.5pt]
                        \multicolumn{1}{c}{Trajectory Error $\downarrow$} & VINS \cite{vins-mono} & VINS(Ours) & ORB-SLAM3 \cite{orb-slam3} \\
                        \midrule
                        two-floors-csc1 & 8.97 & \textbf{2.96} & lost \\
                        two-floors-csc2 & 17.81 & \textbf{10.67} & lost \\ 
                        \bottomrule[1pt]
                \end{tabular}}
        \end{center}
\end{table}

\begin{table}[!h]
        \begin{center}
                \caption{comparison of trajectory errors on the EuRoC dataset}
                \label{te5}
                \setlength{\tabcolsep}{1.5mm}{
                \begin{tabular}{ccccccc}
                        \toprule[1pt]
                        % \midrule[0.5pt]
                        Sequence & Method  & $APE_{rot}$ & $APE_{trans}$ & $RPE_{rot}$ & $RPE_{trans}$\\
                        \midrule
                        \multirow{2}{*}{V1\_02} & Ours & \textbf{3.65} & 0.16 & \textbf{0.18} & 5.0e-2\\
                        & VINS & 3.73 & \textbf{0.12} & 0.22 & \textbf{4.5e-2} \\
                        \midrule
                        \multirow{2}{*}{V1\_03}& Ours & \textbf{2.48} & \textbf{0.08} & \textbf{0.21} & \textbf{5.7e-2} \\
                        &  VINS & 5.56 & 0.15 & 0.25 & 6.9e-2 \\
                        \midrule
                        \multirow{2}{*}{V1\_04} & Ours & \textbf{3.36} & \textbf{0.13} & 0.19 & 4.0e-2 \\
                        & VINS & 3.65 & 0.16 & \textbf{0.18} & \textbf{3.9e-2} \\
                        \midrule
                        \multirow{2}{*}{MH\_03} & Ours & \textbf{1.48} & \textbf{0.12} & \textbf{0.10} & 6.8e-2 \\
                        &  VINS & 1.69 & 0.19 & 0.11 & \textbf{5.9e-2} \\
                        \midrule
                        \multirow{2}{*}{MH\_04} &Ours & \textbf{1.80} & \textbf{0.15} & \textbf{0.14} & 6.3e-2 \\
                        & VINS & 1.87 & 0.18 & 0.16 & \textbf{6.2e-2} \\
                        \bottomrule[1pt]
                \end{tabular}}
        \end{center}
\end{table} 

\subsection{Ablation experiment}
This subsection contains two comparison experiments that validate the line peaky loss and the mNRE loss, respectively. The keypoint repeatability obtained after training with peak loss and line peak loss is shown in table \ref{te7}. During the repeatability calculations NMS radius was $2$ and the number of keypoints was $400$ and performed on the original image size. Similarly, the percentage of correct tracking obtained by training the illumination invariant feature maps using mNRE loss and NRE loss, respectively, is shown in table \ref{te6}. It can be seen that mNRE can effectively improve the quality of optical flow tracking. The experimental parameters are set up in the same way as \ref{kp_tr}.

\begin{table}[!t]
        \begin{center}
                \caption{repeatability in hpatches}
                \label{te7}
                \setlength{\tabcolsep}{3mm}{
                \begin{tabular}{cccccc}
                        \toprule[1pt]
                        Training steps & 10 & 20 & 30 & 40 & 50  \\ 
                        \midrule
                        peaky & 0.20 & 0.22 & 0.23 & 0.22 & 0.23 \\ 
                        line peaky & \textbf{0.32} & \textbf{0.36} & \textbf{0.37} & \textbf{0.39} & \textbf{0.40}\\ 
                        \bottomrule[1pt]
                \end{tabular}}
        \end{center}
\end{table}

\begin{table}[!t]
        \begin{center}
                \caption{correct tracking ratio}
                \label{te6}
                \setlength{\tabcolsep}{2mm}{
                \begin{tabular}{ccccc}
                        \toprule[1pt]
                        Scenes & Indoor & Lighting Source & Outdoor & Image Blur  \\ 
                        \midrule
                        NRE & 0.92 & 0.84 & 0.68 & 0.44 \\ 
                        mNRE & \textbf{0.96} & \textbf{0.91} & \textbf{0.71} & \textbf{0.74} \\ 
                        \bottomrule[1pt]
                \end{tabular}}
        \end{center}
\end{table}

\section{CONCLUSIONS}

% 总结 
% 我们提出了一种混合稀疏光流法，在保持了传统光流法的实时性的同时，利用深度学习的方法提高了其在动态光照场景下的鲁棒性。这项工作的基本思想在于认为卷积网络适合提取图像特征，而传统的LK光流法适合进行光流计算。通过二者的结合共同提高了光流法的性能。为了实现这一目标，我们提出了轻量级网络用于提取关键点和光照不变特征图。并提出了由深层网络辅助所提出的浅层网络训练的流程，以及多个损失函数用于训练网络。在HPatches数据集上验证了所提出的方法的角点重复率，多组动态光照数据集中验证了其在动态光照场景下的性能，最后将其嵌入至VIO系统中，验证了其在实际应用中的有效性。这一工作将为后续视觉SLAM等领域的研究提供新的思路。

In this paper, we propose a hybrid sparse optical flow method that maintains the real-time performance of traditional optical flow method while improving its robustness in dynamic lighting scenes. The basic idea of this work is that CNN are suitable for extracting image features, while traditional LK optical flow method is suitable for optical flow calculation. The combination of the two methods improves the performance of the optical flow method. To achieve this goal, we propose a lightweight network for extracting keypoints and an illumination-invariant feature map. We also propose a training process assisted by a deep network for the shallow network proposed, and multiple loss functions for training the network. The repeatability of the proposed method is verified on the HPatches dataset, and its performance in dynamic lighting scenes is verified on multiple dynamic lighting datasets. Finally, it is embedded in the VIO system to verify its effectiveness in practical applications. This work will support the development of illumination-robust visual SLAM with the hope of achieving robust performance in challenging environments such as caves and tunnels.

\bibliographystyle{IEEEtran}
\bibliography{IEEEabrv,bibtex/IEEEabrv.bib}
\vfill

\end{document}